%% file: preprint.tex
\documentclass[10pt,twocolumn,letterpaper]{article}

 \usepackage[pagenumbers]{cvpr} 

\input{preamble}
\definecolor{cvprblue}{rgb}{0.21,0.49,0.74}
\usepackage[pagebackref,breaklinks,colorlinks,allcolors=cvprblue]{hyperref}

\def\paperID{8} 
\def\confName{CVPR}
\def\confYear{2025}

\title{WildlifeReID-10k: Wildlife re-identification dataset with 10k individual animals}

\author{Lukáš Adam$^1$, Vojtěch Čermák$^2$,  Kostas Papafitsoros$^3$, and Lukas Picek$^{1,4}$ \\
  $^1$UWB in Pilsen, $^2$CTU in Prague, $^3$Queen Mary University of London, $^4$Inria, \\
{\tt\small lukas.adam.cr@gmail.com}, {\tt\small cermavo3@fel.cvut.cz}, {\tt\small k.papafitsoros@qmul.ac.uk}
}

\begin{document}

\twocolumn[{%
\renewcommand\twocolumn[1][]{#1}%
\maketitle
\begin{center}
\vspace{-0.35cm}

    \includegraphics[width=0.95\linewidth]{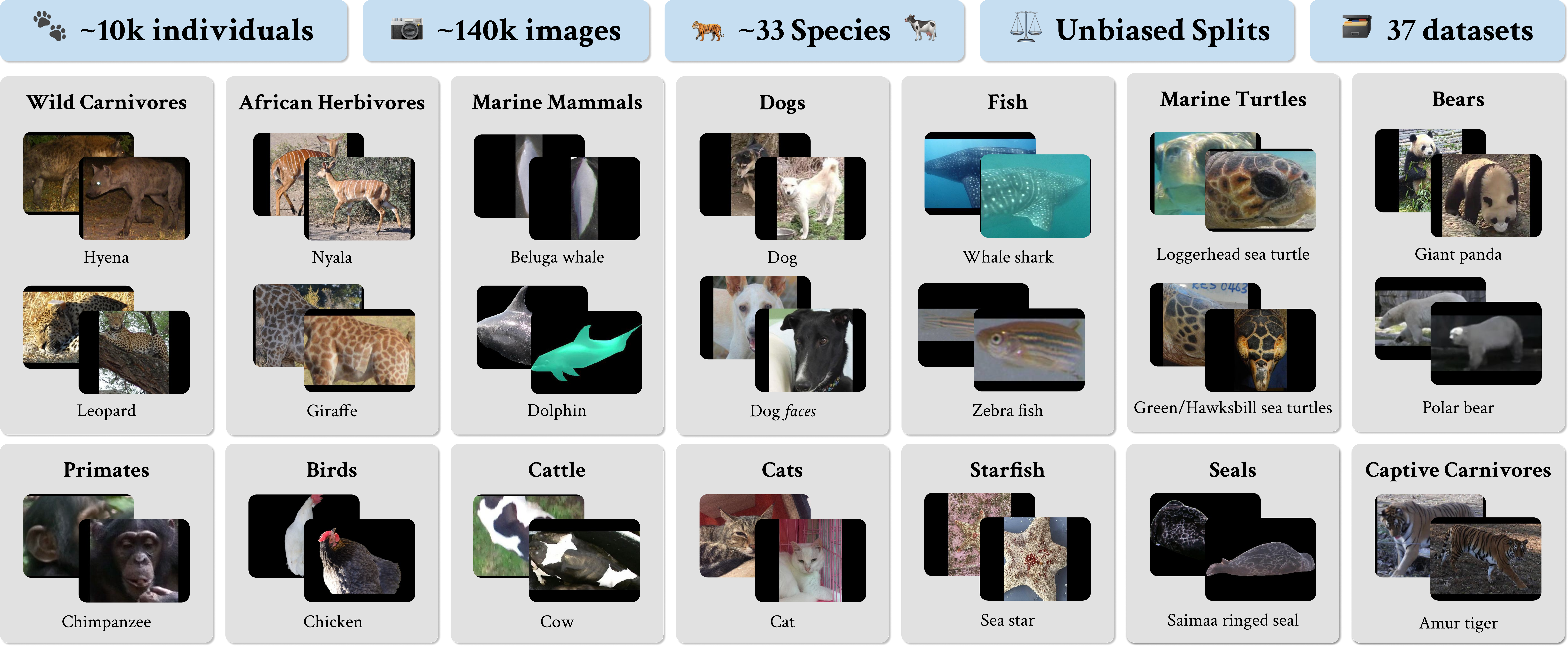}

\end{center}
\vspace{0.5cm}
}]

\maketitle

\begin{abstract}
This paper introduces WildlifeReID-10k, a new large-scale re-identification benchmark with more than 10k animal identities of around 33 species across more than 140k images, re-sampled from 37 existing datasets. WildlifeReID-10k covers diverse animal species and poses significant challenges for SoTA methods, ensuring fair and robust evaluation through its time-aware and similarity-aware split protocol. The latter is designed to address the common issue of training-to-test data leakage caused by visually similar images appearing in both training and test sets.
The \href{https://www.kaggle.com/datasets/wildlifedatasets/wildlifereid-10k}{WildlifeReID-10k} dataset and benchmark are publicly available on Kaggle, along with strong baselines for both closed-set and open-set evaluation, enabling fair, transparent, and standardized evaluation of not just multi-species animal re-identification models.
\end{abstract}

\input{sections/introduction}

\input{sections/related_work}
\input{sections/dataset}
\input{sections/benchmarks}

\input{sections/conclusion}

{
    \small
    \bibliographystyle{ieeenat_fullname}
    \bibliography{main}
}


\end{document}

%% file: preamble.tex
\usepackage[utf8]{inputenc} 
\usepackage[T1]{fontenc}    
\usepackage{booktabs}       
\usepackage{amsfonts}       
\usepackage{nicefrac}       
\usepackage{microtype}      
\usepackage{xcolor}         
\usepackage{comment}
\usepackage{amsmath}
\usepackage{todonotes}
\usepackage{algorithm}
\usepackage{algorithmicx}
\usepackage{algpseudocode}
\usepackage{multirow}
\usepackage{graphicx}
\usepackage{pgfplots}
\usepackage{pgfplotstable}
\usepackage{wrapfig}
\usepackage{caption}
\usepackage{rotating}
\usepgfplotslibrary{groupplots}

\usepackage{pifont}
\newcommand{\cmark}{\ding{51}}
\newcommand{\xmark}{\ding{55}}

\pgfplotstableread[col sep = comma]{csv/fp_tp.csv}\tabFPTP
\pgfplotstableread[col sep = comma]{csv/identity_counts.csv}\tabCounts
\pgfplotstableread[col sep = comma]{csv/normalized.csv}\tabAccuracy
\pgfplotstableread[col sep = comma]{csv/similarity-aware.csv}\tabSimilarity
\pgfplotstableread[col sep = comma]{csv/time-aware.csv}\tabTime

\tikzset{
    lineA/.style= {very thick, blue},
    lineB/.style= {very thick, orange},
    lineC/.style= {very thick, blue, dotted},
    lineD/.style= {very thick, orange, dotted},
}

%% file: sections/introduction.tex
\section{Introduction}
Individual animal re-identification assigns identities to individual animals in images. It is typically based on visual analysis of morphological characteristics, such as marks, spots, and stripes, which are stable over time and unique to each individual. When combined with metadata (e.g., capture time and date, and location), this information helps study various aspects of wild animal populations' biology and ecology. It can support tasks such as disease monitoring and control \cite{palencia2023not}, assessing an animal's role in the ecosystem \cite{rowcliffe2008estimating}, tracking invasive species \cite{caravaggi2016invasive}, and evaluating human impact on habitats and ecological restoration efforts \cite{blount2021covid}.

\begin{figure*}[!ht]
\centering
\includegraphics[width=0.95\textwidth]{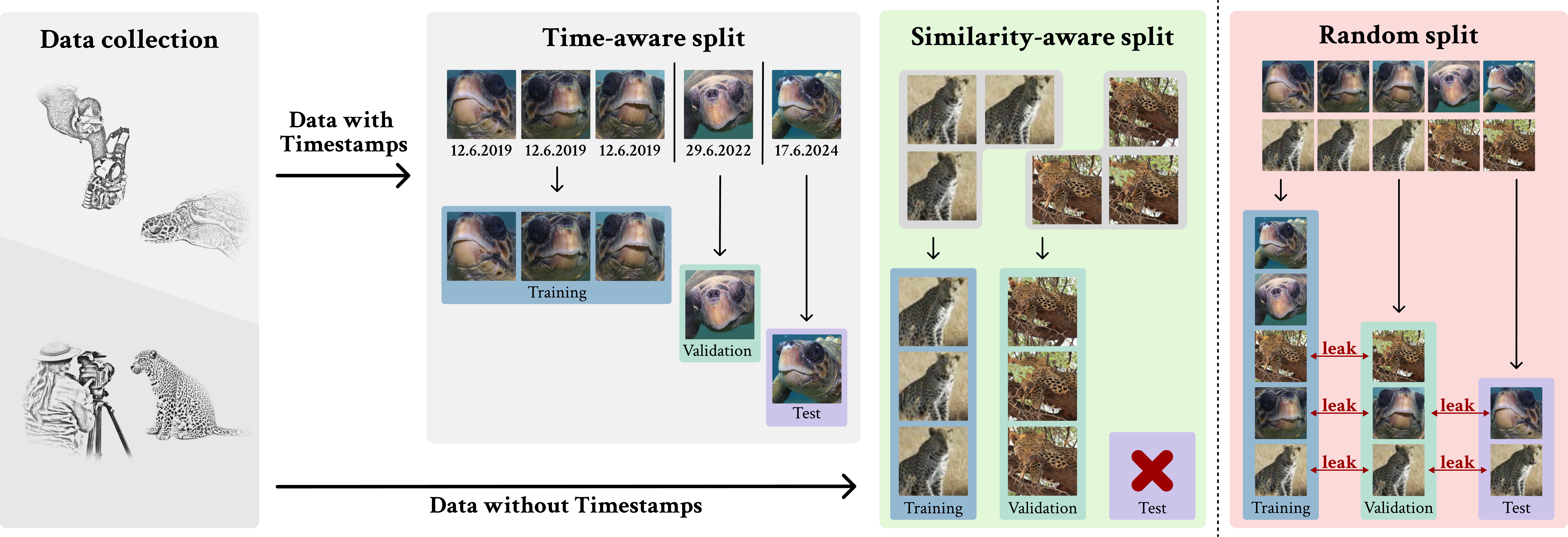}
\caption{\textbf{WildlifeReID-10k splitting methodology.} Data collected in a single encounter is usually split randomly and might be \textit{corrupted} by unwanted training-to-test set leakage.
Therefore, we employ \textbf{(i) a time-aware split} if timestamps are available, and \textbf{(ii) a similarity-aware split}, where visually similar images are treated as a single observation which is assigned to either the training, validation, or test set.}
\label{fig:overview}
\vspace{-0.25cm}
\end{figure*}

In recent years, there has been a significant increase in research on automatic re-identification methods for wild animals, resulting in the first-ever foundational models such as MegaDescriptor \cite{Cermak_2024_WACV}, MiewID \cite{otarashvili2024multispecies}, and WildFusion \cite{cermak2024wildfusion}. This surge has been made possible by recent advances in computer vision and machine learning and has also been driven by the growing availability of wildlife imagery obtained through various sources such as camera traps \cite{schneider2019past}, drones \cite{landeo2020using}, satellites \cite{wall2014novel}, regular photo-identification surveys \cite{Schofield_2020}, and passive crowdsourcing from social media data \cite{papafitsoros2023social}. This increase in research is evidenced not only by the increasing number of research studies \cite{bedetti2020system,bruslund2020re,reno2019sift} and the corresponding review papers \cite{schneider2022similarity,vidal2021perspectives,ravoor2020deep,schneider2019past} but also by the expanding number of publicly available datasets for animal re-identification that cover numerous animal groups, including primates\,\,\cite{freytag2016chimpanzee,witham2018automated}, carnivores\,\,\cite{li2019atrw,botswana2022,dlamini2020automated}, reptiles\,\,\cite{zinditurtles, dunbar2021hotspotter}, whales\,\,\cite{cheeseman2022advanced}, or mammals\,\,\cite{belugaid,trotter2020ndd20,zuffi2019three}.

However, many datasets have several limitations. Due to the high cost of labeling, most contain only a limited number of photos and individuals, reducing their standalone value. Documentation is often lacking, and formats are rarely unified, making initial analysis time-consuming. Even more critically, standardised training-testing splits and evaluation metrics are typically absent. As a result, numerous methods were often evaluated on only a subset of available datasets, with limited comparison to prior work \cite{jiao2024toward,schneider2022similarity,shukla2019hybrid}. 

Some of these issues were addressed by Cermak et al. \cite{Cermak_2024_WACV}, who released the \href{https://github.com/WildlifeDatasets/wildlife-datasets}{\texttt{wildlife-datasets}} library, which greatly simplifies dataset accessibility and handling. However, the library did not consider the "\textit{natural}" open-set scenario where new identities are introduced, and does not address the training-to-test dataset leakage issues raised by Adam et al. \cite{adam2024seaturtleid}. Specifically, it is argued  that the standard evaluation protocol, which randomly splits the dataset into training and testing sets, results in information leakage between the two. This leakage inflates performance metrics, as animal re-identification datasets often contain multiple similar photos taken during a single human-animal encounter\footnote{Usually one observation of an individual from one location and time.} or extracted from consecutive video frames.

Therefore, while constructing WildlifeReID-10k, we employ a \textbf{time-aware split} \cite{adam2024seaturtleid} whenever timestamps are available, to prevent this leakage. More specifically, we select cutoff dates for each identity and assign all images captured before (after) this date to the training  (validation or test) set. This guarantees that photos from the same encounter are placed either in the training or test set, but not both. Since most datasets do not have images with timestamps, we propose a new approach for encounter estimation by \textbf{visual similarity clustering}, where each cluster is treated as an observation, and all of its images are assigned to either the training or test set. See \cref{fig:overview} for a visual explanation.

WildlifeReID-10k is a curated animal re-identification benchmark that enables fair evaluation of models. Its relation to other existing dataset collections, such as PetFace \cite{shinoda2024petface}, Wildlife-71 \cite{jiao2024toward}, or any private dataset collection, is complementary, since these collections can serve as additional data sources for training models. However, we recommend using WildlifeReID-10k for a fair comparison of model performance, which we hope will facilitate an unbiased measure of progress in animal re-identification.

\vspace{0.25cm}
\noindent\textbf{The main contributions of this paper include:}

\begin{itemize}
    \item Splitting and evaluation methodology that prevents training-to-test data leakage, and allows fair and \textit{less biased} evaluation without inflated performance.
    \item New dataset and benchmark -- \href{www.kaggle.com/datasets/wildlifedatasets/wildlifereid-10k}{WildlifeReID-10k} -- for animal re-identification with more than 10k identities and 140k images of around 33 species.
    \item A set of baselines for both closed- and open-set settings.
\end{itemize}

%% file: sections/related_work.tex
\section{Related work}

\noindent\textbf{Datasets:} Publicly available datasets and tools have accelerated efforts to standardize and benchmark animal re-identification. A key contribution was made by Cermak et al. \cite{Cermak_2024_WACV}, who provided a thorough overview of existing datasets and introduced a unifying library (\href{https://github.com/WildlifeDatasets/wildlife-datasets}{\texttt{wildlife-datasets}}) that simplifies the dataset loading, preprocessing, and evaluation process, enabling fast, replicable, and comparable experiments across many species and scenarios. Following its success, several works introduced new datasets.

Jiao et al. introduced the Wildlife-71 dataset \cite{jiao2024toward}, using three existing datasets -- ATRW \cite{li2019atrw}, GiraffeZebraID \cite{parham2017animal}, and SealID \cite{nepovinnykh2022sealid} -- alongside manually labeled identities from the GOT-10k tracking dataset \cite{huang2019got} and 816 YouTube videos. A significant part of Wildlife-71 is sourced from video datasets, which reduces the diversity of the images.

Shinoda et al. collected animal photos from pet shops and adoption websites and created the PetFace dataset \cite{shinoda2024petface}. Such an approach allows fast data collection of a large number of individuals, but raises potential ethical and privacy concerns. Moreover, over 90\% of images are from indoors, i.e., not captured in the wild. \vspace{-0.1cm} \\

\noindent\textbf{Methods:} 
Most animal re-identification methods are developed and evaluated under a closed-set setting, where the identities present during testing are only those that have been included in the training set \cite{Cermak_2024_WACV,cermak2024wildfusion,li2019atrw,otarashvili2024multispecies}. This set-up simplifies the problem but fails to reflect real-world conditions, where animal populations are dynamic and new individuals are entering or leaving populations due to migration, birth, or other factors. As a result, models trained in a closed-set setting, despite their great performance, are limited in their ability to generalize to these unseen individuals.

Open-set recognition works under a more realistic setting as new identities are expected in the test set. Existing methods typically assign novelty scores based on softmax confidence, distance to decision boundaries, or embedding space proximity \cite{hendrycks2017a,vaze2022openset,vojivr2024pixood}. While widely used in other domains, this setting is still overlooked in animal re-identification. This gap highlights the need for more research into methods that generalize to novel individuals and support robust evaluation in open-set scenarios.




%% file: sections/dataset.tex
\section{WildlifeReID-10k}

The construction of the WildlifeReID-10k is built around the \href{https://github.com/WildlifeDatasets/wildlife-datasets}{\texttt{wildlife-datasets}} \cite{Cermak_2024_WACV} framework -- that unifies available wildlife re-identification datasets and provides an API that allows their straightforward access and download. WildlifeReID-10k includes 24 available datasets (bottom of \cref{table:datasets}) and extends the framework with 13 new datasets\footnote{All selected datasets have permissible licenses for being re-uploaded.} (top of \cref{table:datasets}) that were not available earlier.
In total, the dataset contains 140,488 images and 10,772 individuals of around 33 species. All images were either pre-cropped or cropped using the provided bounding boxes or segmentation masks, resulting in a dataset size of only 26 GB -- or approximately 185 kB per image on average -- without any resizing.

\begin{table}[!hb]
\vspace{-0.1cm}
\small
\centering
\caption{\textbf{Datasets used to create WildlifeReID-10k.} The (top) section lists newly added datasets, while the (bottom) section lists previously available datasets integrated into the collection.}
\vspace{-0.1cm}
\label{table:datasets}
\begin{tabular}{@{}l@{\hspace{-15px}}r@{\hspace{5px}}r@{\hspace{8px}}r@{\hspace{8px}}r@{}}
\toprule
 & \textbf{timestamp} & \textbf{images} & \textbf{identities} & \textbf{ratio} \\
\midrule
AmvrakikosTurtles \cite{adam2024exploiting} & \cmark & 200 & 50 & 4.0 \\
CatIndividualImages \cite{catindividuals} & \xmark & 13,021 & 509 & 25.6 \\
Chicks4FreeID \cite{kern2024towards} & \xmark & 1,146 & 50 & 22.9 \\
CowDataset \cite{cowdataset} & \cmark & 1,485 & 13 & 114.2 \\
DogFaceNet \cite{mougeot2019deep} & \xmark & 8,363 & 1,393 & 6.0 \\
NDD20 \cite{trotter2020ndd20} & \xmark & 2,657 & 82 & 32.4 \\
MultiCamCows2024 \cite{yu2024multicamcows2024} & \cmark & 5,112 & 90 & 56.8 \\
PolarBearVidID \cite{zuerl2023polarbearvidid} & \xmark & 1,391 & 13 & 107.0 \\
PrimFace \cite{primface} & \xmark & 1,282 & 68 & 18.9 \\
ReunionTurtles \cite{adam2024exploiting} & \cmark & 336 & 84 & 4.0 \\
SeaStarReID2023 \cite{wahltinez2024open} & \cmark & 2,187 & 95 & 23.0 \\
SouthernProvinceTurtles \cite{adam2024exploiting} & \xmark & 481 & 51 & 9.4 \\
ZakynthosTurtles \cite{adam2024exploiting} & \cmark & 160 & 40 & 4.0 \\
\midrule
AAUZebraFish \cite{bruslund2020re} & \xmark & 336 & 6 & 56.0 \\
AerialCattle2017 \cite{andrew2017visual} & \xmark & 2,329 & 23 & 101.3 \\
ATRW \cite{li2019atrw} & \xmark & 5,415 & 182 & 29.8 \\
BelugaID \cite{belugaid} & \cmark & 8,559 & 788 & 10.9 \\
BirdIndividualID \cite{ferreira2020deep} & \cmark & 2,629 & 50 & 52.6 \\
Cows2021 \cite{gao2021towards} & \cmark & 8,670 & 179 & 48.4 \\
CTai \cite{freytag2016chimpanzee} & \xmark & 4,662 & 71 & 65.7 \\
CZoo \cite{freytag2016chimpanzee} & \xmark & 2,109 & 24 & 87.9 \\
FriesianCattle2015 \cite{andrew2016automatic} & \xmark & 193 & 25 & 7.7 \\
FriesianCattle2017 \cite{andrew2017visual} & \xmark & 940 & 89 & 10.6 \\
Giraffes \cite{miele2021revisiting} & \cmark & 1,368 & 178 & 7.7 \\
GiraffeZebraID \cite{parham2017animal} & \cmark & 6,898 & 2,051 & 3.4 \\
HyenaID2022 \cite{botswana2022} & \xmark & 3,129 & 256 & 12.2 \\
IPanda50 \cite{wang2021giant} & \xmark & 6,874 & 50 & 137.5 \\
LeopardID2022 \cite{botswana2022} & \xmark & 6,806 & 430 & 15.8 \\
MPDD \cite{he2023animal} & \xmark & 1,657 & 191 & 8.7 \\
NyalaData \cite{dlamini2020automated} & \xmark & 1,942 & 237 & 8.2 \\
OpenCows2020 \cite{andrew2021visual} & \xmark & 4,736 & 46 & 103.0 \\
SealID \cite{nepovinnykh2022sealid} & \xmark & 2,080 & 57 & 36.5 \\
SeaTurtleID2022 \cite{adam2024seaturtleid} & \cmark & 8,729 & 438 & 19.9 \\
SMALST \cite{zuffi2019three} & \xmark & 1,290 & 10 & 129.0 \\
StripeSpotter \cite{lahiri2011biometric} & \cmark & 820 & 45 & 18.2 \\
WhaleSharkID \cite{holmberg2009estimating} & \xmark & 7,693 & 543 & 14.2 \\
ZindiTurtleRecall \cite{zinditurtles} & \xmark & 12,803 & 2,265 & 5.7 \\
\midrule
TOTAL & 13 & 140,488 & 10,772 & 13.0 \\
\bottomrule
\end{tabular}
\end{table}

To further improve dataset quality, i.e., to prevent image repetition and to balance the ratio of images per individual, we selected every $k$-th frame from several datasets (\textit{AAUZebraFish}, \textit{AerialCattle2017}, \textit{BirdIndividualID}, \textit{MultiCamCows2024} $k=20$; \textit{PolarBearVidID} $k=100$; \textit{SMALST} $k=10$). Without this, for instance \textit{PolarBearVidID} would contain 140k images of just 13 individuals, and WildlifeReID-10k would be  dominated by polar bears.

Additionally, we updated two datasets with new data (\textit{BelugaID} and \textit{SeaTurtleID2022}) and fixed several wrong labels in 3 additional datasets (\textit{CTai}, \textit{Cows2021}, \textit{FriesianCattle2015}). In all datasets, we also found unidentified animals (being named ``Adult'', ``new\_whale'' or ``\_\_\_\_'') which were also excluded. Consequently, around half of the data in WildlifeReID-10k ($18/37$ datasets) are either brand new or updated. Detailed scripts used to create WildlifeReID-10k are available through the \href{https://github.com/WildlifeDatasets/wildlife-datasets}{\texttt{wildlife-datasets}}.

While there are several datasets with domestic animals, most of the animals are wild. Indeed, \cref{table:datasets} implies that 50\% of individuals were observed in the wild (71\% when including rehabilitation centres). This number excludes wild animals in the zoos and confirms that the WildlifeReID-10k dataset indeed focuses mostly on wild animals.

\subsection{Construction of default splits}

The inconsistent training/test data splitting and evaluation methodology is a significant limitation to further animal re-identification research. A commonly used approach is the closed-set setting, where all individuals present in the test set have already been observed during training. In this scenario, each new image is of the known identities, and the assumption is that no other individuals will appear. This suits controlled environments, such as farms or zoos, where the population is fully known and static. However, this assumption rarely holds in wildlife monitoring. In contrast, the open-set setting acknowledges that new data may include previously unobserved individuals, such as newly born animals or individuals not yet included in the dataset.

Another methodological issue arises from the way that training and test sets are constructed. Developers often split images randomly without accounting for the structure of wildlife data. These datasets typically contain visually similar images, e.g., multiple photos taken during a human-animal encounter or consecutive frames extracted from videos. Random splitting can result in similar training and test images, leading to train-test leakage \cite{adam2024seaturtleid}. This leakage makes the classification task artificially simplified and can severely inflate reported performance, undermining the real-world utility of the proposed methods.

To the best of our knowledge, all available datasets either have no default split or use a random split without considering data acquisition details with only one exception (i.e., \textit{SeaTurtleID2022}), which splits data based on timestamps -- images taken before a specific date are assigned to the training set and those after are reserved for the test set. Hence, we propose a method for generating splits to mitigate training-to-test leakage. The method work as follows:\vspace{-0.15cm}\\

\noindent\textbf{Step\,\,1: Select open-set identities}: To allow both open- and closed-set evaluation, we first select around 5\% of all identities from each dataset, and we put all their images directly into the open-set test set. \vspace{-0.15cm}\\

\noindent\textbf{Step\,\,2: Time-aware clustering}: From the remaining 95\% we check whether the dataset contains timestamps, and if so, we split it in a time-aware fashion. Specifically, for each individual, we selected approximately 85\% of the oldest observations and put them in the training set and the remaining into the test set. From datasets with timestamps (see \cref{table:datasets}), we excluded \textit{AmvrakikosTurtles}, \textit{ReunionTurtles}, \textit{StripeSpotter}, and \textit{ZakynthosTurtles} due to a limited number of images or encounters. \vspace{-0.15cm}\\

\noindent\textbf{Step\,\,3: Similarity-aware clustering}:
Most datasets lack timestamp metadata, making it difficult to determine encounters. Since images from the same encounter are often near-duplicates, they should lie close in the feature space of foundational models, such as DINOv2 \cite{oquab2023dinov2} or CLIP \cite{radford2021learning}.

Therefore, we first feed-forward all the data into the DINOv2 and store the resulting embeddings. To identify encounters, we use a variant of the single-linkage hierarchical clustering algorithm with a stopping criterion based on cosine similarity threshold $\theta$; in our case, set manually for each dataset to optimize the number of correct clusters (see Appendix \ref{apx:quality_of_clusters} for more details). For each identity, we construct a graph where each image is a node, and an edge connects two nodes if their similarity is above the threshold. The connected components in this graph form our clusters. A more precise description can be found in \cref{algo:clustering}.

These clusters are then used to create the training-test split. For each identity, all its clusters are placed into the training set. The unclustered images are randomly divided into the training and test sets so that the training set contains the desired number of samples.

\begin{algorithm}[!hb]
\small
\caption{for finding clusters for the training-test split}
\label{algo:clustering}
\begin{algorithmic}[1]
\Require $S \in \mathbb{R}^{n \times n}$ similarity matrix, $\theta$ threshold
\State $\text{parent} \gets [0, 1, \ldots, n-1]$
\State Build and sort edges $(i,j,S_{i,j})$ by descending similarity
\Function{Find}{$i$}
    \While{$\text{parent}[i] \neq i$}
        \State $\text{parent}[i] \gets \text{parent}[\text{parent}[i]]$
        \State $i \gets \text{parent}[i]$
    \EndWhile
    \State \Return $i$
\EndFunction
\For{$(i, j, \text{sim})$ \textbf{in} sorted edges}
    \If{$\text{sim} < \theta$} \textbf{break} \EndIf
    \If{$\text{Find}(i) \neq \text{Find}(j)$}
        \State $\text{parent}[\text{Find}(i)] \gets \text{Find}(j)$ \Comment{merge clusters}
    \EndIf
\EndFor
\State Build clusters by grouping nodes with same $\text{Find}(i)$ root
\State \Return clusters
\end{algorithmic}
\end{algorithm}

\vspace{-4mm}
\subsection{Dataset statistics}\label{sec:statistics}
Following the above-mentioned approach, we formed two splits that are further described in \cref{table:splits}. The closed-set setting contains 109,927 training images and 22,745 test images, with all test identities seen during training. In the open-set split, 523 new identities appear in the test set, resulting in 30,561 test images.

\begin{table}[!h]
\setlength{\tabcolsep}{5pt}
\small
\centering
\caption{\textbf{WildlifeReID-10k -- proposed splits.} Values are grouped to highlight contrasts between closed- and open-set settings.}
\vspace{-0.1cm}
\label{table:splits}
\begin{tabular}{@{}lcccccc@{}}
\toprule
 & \multicolumn{2}{c}{\textbf{Images}} & \multicolumn{3}{c@{}}{\textbf{Identities}} \\
\cmidrule(lr){2-3} \cmidrule(lr){4-6}
\textbf{Setting} & Training & Test & Training & Test$_\text{\textit{known}}$ & Test$_\text{\textit{new}}$ \\
\midrule
\textit{Closed-set} & \multirow{2}{*}{109,927} & 22,745 & \multirow{2}{*}{10,249} & \multirow{2}{*}{8,391} & -- \\
\textit{Open-set}   &                          & 30,561 &                          &  & 523 \\
\bottomrule
\end{tabular}
\end{table}

Approximately 56\% of individuals in the test set have only one image, while the top 1\% account for 19\% of all images and the top 10\% for 53\%. This highlights a strong imbalance typical of datasets collected in the wild, where certain animals are captured more frequently -- often because they frequent high-traffic areas like camera trap routes or shallow coastal waters. As a result, standard top-k accuracy metrics can be misleading, disproportionately favoring frequently observed and often well-studied individuals. To address this, we also report balanced accuracy, which accounts for such skew. In \cref{fig:identity_counts}, we illustrate the distribution of images per individual in the training and test sets.

\begin{figure}[!t]
\centering
\begin{tikzpicture}
\begin{axis}[
    height=2.5cm,
    width=7cm,
    scale only axis,
    ylabel style={
        yshift=-3mm
    },
    xticklabel style={
        /pgf/number format/fixed,
        /pgf/number format/precision=3,
    },
    xmin=0,
    xmax=10500,
    scaled ticks=false,
    xlabel={individual (class)},
    ylabel={number of images},
    grid=major,
    ytick={0,1,2},
    yticklabels={1,10,100}
]
\addplot [smooth, very thick, blue] table[x index=0, y index=1] {\tabCounts}; \addlegendentry{training}
\addplot [smooth, very thick, red] table[x index=0, y index=2] {\tabCounts}; \addlegendentry{test}
\end{axis}
\end{tikzpicture}
\caption{\textbf{Number\,of\,images per individual\,in WildlifeReID-10k}.} 
\label{fig:identity_counts}
\vspace{-4mm}
\end{figure}

\subsection{Quality of clusters}
\label{apx:quality_of_clusters}
We can verify the clustering quality (i.e., purity) on datasets with timestamps. For each dataset, we design the \textit{true-positives} \text{TP} and \textit{false-positives} \text{FP} counts by:
\begin{equation}\label{eq:tpfp}
\text{TP} = \sum_{c\in C}(|c| - 1), \hspace{4mm}
\text{FP} = \sum_{c\in C}(|c| - n_{\text{max date}}(c)),
\end{equation}
where $C$ is the set of clusters. TP measures how many images were clustered above trivial singleton clusters. FP subtracts the number of images with the most frequent date from the cluster size.
The clustering threshold $\theta$ controls a trade-off between cluster size and its quality. 

On the one hand, the algorithm aims for large clusters per identity to address near duplicates in the dataset, which requires maximizing true positives (TP). On the other hand, overly large clusters may group images from different encounters, introducing noise.
This shifts the task away from original identification by clustering not just duplicates but also images with true similarity features, which artificially increases the difficulty.
Thus, false positives (FP) reflect cluster impurity and should be minimized.

In Figure \ref{fig:clusters} we show the dependence of TP and FP on the parameter $\theta$. When $\theta$ is large, no clusters are formed, and both counts are $0$. On the other hand, when $\theta$ is small, all images belong to one cluster, and both counts are large. 

\begin{figure}[!t]
\centering
\begin{tikzpicture}
\begin{axis}[
    height=2.5cm,
    width=6.25cm,
    scale only axis,
    ylabel style={
        yshift=3mm
    },
    xticklabel style={
        /pgf/number format/fixed,
        /pgf/number format/precision=3,
    },
    xmin=0.9,
    xmax=1.0,
    scaled ticks=false,
    xlabel={threshold $\theta$},
    ylabel={number of images},
    grid=major,
]
\addplot [smooth, very thick, blue] table[x index=0, y index=1] {\tabFPTP}; \addlegendentry{{\small cluster impurity (FP)}}
\addplot [smooth, very thick, red] table[x index=0, y index=2] {\tabFPTP}; \addlegendentry{{\small clustered images (TP)}}
\end{axis}
\end{tikzpicture}
\caption{\textbf{Size of clusters and their impurity based on $\theta$}.}
\label{fig:clusters}
\vspace{-4mm}
\end{figure}

An optimal threshold keeps false positives (FP) low. In our case, with $\theta = 0.97$ we found 5,601 duplicates (TP) out of 45,637 images, with 685 incorrectly grouped (FP). However, since the datasets differ, using just one threshold not ideal. 
To better account for dataset variability, one could manually select a threshold for each dataset based on visual inspection and improve the results significantly, i.e., got 8,206 TPs at 464 FPs in our case.
Since the \textit{manual} method was better, we used it for the similarity-aware split when timestamps were unavailable.

\begin{figure}[!b]
\vspace{1mm}
\centering
    \includegraphics[width=0.95\linewidth]{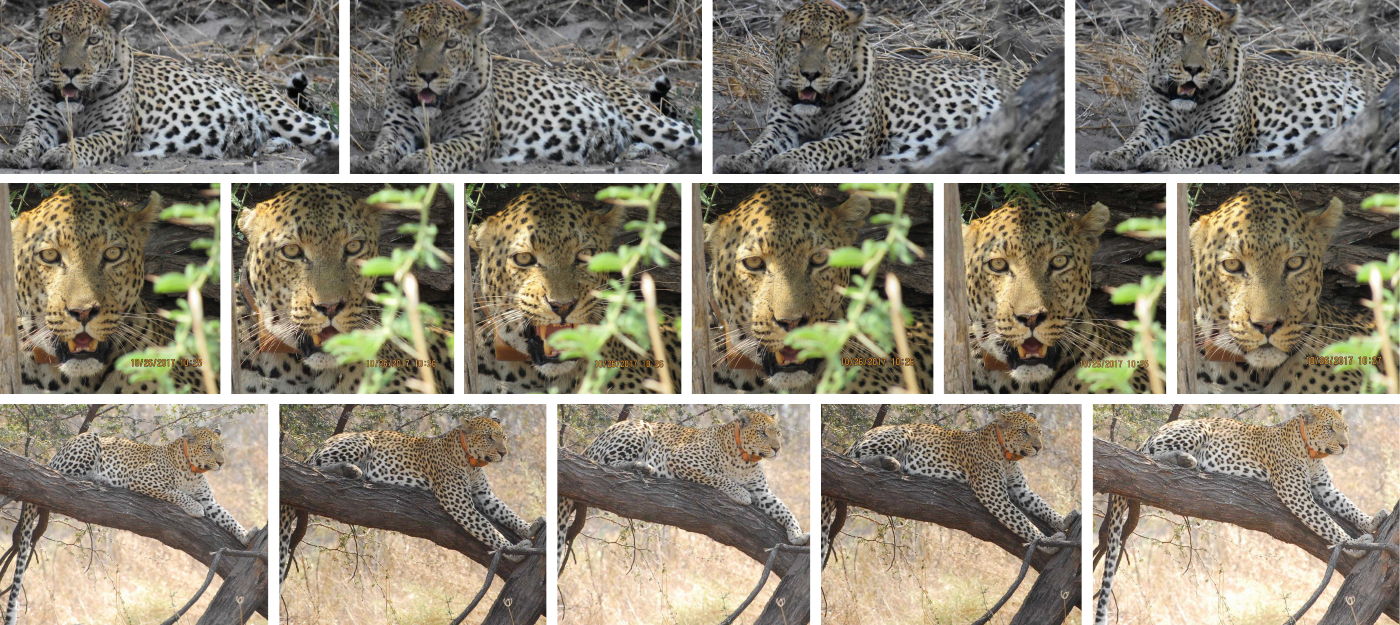}
    \caption{\textbf{Found clusters (columns) in LeopardID2022 dataset}. Most clusters consist of almost identical images with only small differences in size or the leopard's head position.}
\label{fig:split_similar_images}
\end{figure}

Table \ref{table:quality} and Figure \ref{fig:split_similar_images} provide further evidence of the high quality of the obtained clusters. In the Table \ref{table:quality}, rows indicate cluster size, columns the number of unique timestamps, and entries the number of clusters per size-timestamp combination. The final column shows the share of clusters with a single timestamp, indicating high-quality, time-consistent groupings. While small clusters almost always come from a single encounter, even the largest clusters maintain over 77\% consistency.

\begin{table}[!b]
\vspace{-2mm}
\centering
\caption{\textbf{Number of clusters with certain sizes (rows) and unique timestamps (columns) for our clustering algorithm.} The smaller the number of unique timestamps in a cluster, the higher quality of the corresponding cluster.}
\label{table:quality}
\begin{tabular}{@{}lrllllll@{}}
\toprule
\textbf{cluster} & \multicolumn{6}{c}{\textbf{unique timestamps}} & \textbf{single} \\ \cmidrule(lr){2-7}
\textbf{size} & 1 & 2 & 3 & 4 & 5 & 6+ & \textbf{time} \\
\midrule
2 & 3009 & 78 &  &  &  &  & 97.5\% \\
3 & 374 & 27 & 8 &  &  &  & 91.4\% \\
4 & 193 & 26 & 4 & 2 &  &  & 85.8\% \\
5 & 127 & 16 & 2 & 1 & 0 &  & 87.0\% \\
6+ & 212 & 45 & 14 & 2 & 1 & 0 & 77.4\% \\
\bottomrule
\end{tabular}
\end{table}


%% file: sections/benchmarks.tex
\section{Baselines}
This section outlines the evaluation setup for both closed-set and open-set benchmarks. The closed-set setting measures model performance on known individuals and is suitable for controlled environments. The open-set setting reflects real-world dynamics, testing whether models can recognize unseen individuals and distinguish them from known ones.

\subsection{Experimental Settings}
To establish strong and robust baselines, we train models across both CNN and transformer-based architectures. Specifically, we use ConvNeXt-Base and EfficientNet-B3 as CNN models, and ViT-Base/p16 and Swin-Base/p4w7 as transformer-based alternatives. 

While training, we follow the established methodology \cite{Cermak_2024_WACV,otarashvili2024multispecies} and optimize the models using metric learning with ArcFace loss \cite{deng2019arcface} that aims to learn a feature extractor that produces highly discriminative feature embeddings. All models were trained for 100 epochs using SGD with momentum (m=0.9), cosine schedule (10$e^{-2}$ to 10$e^{-4}$ and 10$e^{-3}$ to 10$e^{-6}$ for CNNs and Transformers respectively) and ArcFace with 64 scale and 0.5 margin. 

Additionally, we train the same architectures but with cross-entropy loss. In this case, we used Adam optimizer \cite{kingma2014adam} with cosine schedule from 10$e^{-4}$ to 10$e^{-6}$ for all. \\

\noindent\textbf{Closed-set methodology}.
Models trained with ArcFace are used as feature extractors, and embeddings are computed for both training and test images. A 1-nearest neighbor classifier is constructed using the training embeddings, and test identities are predicted accordingly. For models trained with cross-entropy loss, predictions are made based on the index of the highest logit value. \\

\noindent\textbf{Open-set methodology}.
ArcFace-trained models are evaluated using scores based on feature-space distance. Specifically, the Nearest Mean Score (NMS) \cite{vojivr2024pixood}, which measures the cosine distance to the centroid of the closest class.
For cross-entropy models, the Maximum Softmax Probability (MSP) \cite{hendrycks2017a} and Maximum Logit Score (MLS) \cite{vaze2022openset} are calculated, corresponding to the highest predicted probability and logit value, respectively. \\

\noindent{\textbf{Foundation models for animal re-id:}}
To evaluate the generalization capabilities of existing foundation models for animal re-identification, we benchmark recent state-of-the-art feature extractors. Specifically, we use the pre-trained MiewID-msv3 model \cite{otarashvili2024multispecies}, which is designed for multi-species re-identification. We exclude MegaDescriptor-L \cite{Cermak_2024_WACV} from our evaluation due to a substantial overlap between its training data and our test datasets, which would compromise the fairness of the comparison.

Despite a similar overlap with the training data of MiewID-msv3, we still include it in our experiments to assess its performance in real-world, non-curated scenarios and to evaluate the robustness of our proposed split strategies under such conditions.
We evaluate the model in both \textbf{closed-} and \textbf{open-set} scenarios. In the closed-set setting, identification is done by retrieving the most similar training image based on cosine similarity. In the open-set setting, an image is classified as a new individual if its highest similarity score falls below a predefined threshold (see Eq. \eqref{eq:predict}).

To test the robustness of our proposed split strategy, we also include a \textbf{random split baseline}, ensuring the same number of training and test images per individual. 

\subsection{Metrics}

\noindent\textbf{Closed-set metrics:} Closed-set classification is a well-defined problem, so standard metrics are used: top-1 (mTop1) and top-5 accuracy (mTop5), both macro-averaged across datasets to reflect expected image-level performance. Following \cref{sec:statistics}, we also report balanced top-1 accuracy over datasets and individuals, referred to as BAKS (balanced accuracy on known samples) \cite{mendes2017nearest}. Mathematically:
\begin{equation}
\aligned
\small
\text{mTop1} &= \frac{1}{D}\sum_{d=1}^{D}\frac{1}{|I^{d}|}\sum_{i\in I^{d}} \textbf{1}(\hat y_i = y_i),\\
\text{BAKS} &= \frac{1}{D}\sum_{d=1}^{D}\frac{1}{|C_d|}\sum_{c\in C_d} \frac{1}{|I^{d,c}|}\sum_{i\in I^{d,c}} \textbf{1}(\hat y_i = y_i),
\endaligned
\end{equation}
where $\textbf{1}$ is the indicator (0/1) function, $D$ the number of datasets, $C_d$, the \textit{known} classes (individuals) in dataset $d$ and $I^{d,c}$ (or $I^d$) indices of all images of class $c$ (or all classes) in the test set of the dataset $d$. Finally, $y_i$ and $\hat y_i$ are the true and predicted identities of the $i$-th image. \\

\noindent\textbf{Open-set classification metrics} determine whether the individual is new (detection) or provide an identity if known (classification). The prediction is usually achieved by outputting a class $c$ as well as a confidence score $t$ that indicates how likely it is for the image to have a new identity. Whenever the confidence score is below some prescribed threshold $t_{\rm new}$, the method predict the new class instead of $c$:
\begin{equation}\label{eq:predict}
\text{prediction} = \begin{cases} c &\text{if }t\ge t_{\rm new}, \\ \text{new individual} &\text{otherwise.} \end{cases}
\end{equation}
The threshold $t_{\rm new}$ controls if a new individual is predicted, higher values equal a new individual, and lower values equal none, making open-set classification effectively a closed-set task with an additional class. \\

\noindent\textbf{Open-set detection metrics:} Detection of new individuals is a binary task, with known individuals treated as the positive class. Detection metrics fall into two categories depending on whether they require specifying $t_{\rm new}$. Some methods select this threshold automatically or average over all possible values. For instance, TNR@95TPR finds $t_{\rm new}$ such that at least 95\% of known individuals are correctly identified (true positive rate, TPR), then reports the proportion of new individuals correctly detected (true negative rate, TNR). AUROC, on the other hand, computes the area under the ROC curve, which plots TPR against the false positive rate (FPR); since $\text{FPR}=1-\text{TNR}$, AUROC effectively aggregates TNR@$\alpha$TPR across all $\alpha \in [0,1]$.

Balanced accuracy on unknown samples is an example of metrics requiring the threshold $t_{\rm new}$, defined as
\begin{equation}
\small
\operatorname{BAUS} = \frac{1}{D}\sum_{d=1}^{D}\frac{1}{|C_d^*|}\sum_{c\in C_d^*} \frac{1}{|I^{d,c}|}\sum_{i\in I^{d,c}} \textbf{1}(\hat y_i = \text{new}).
\end{equation}
It is the same as BAKS, but the middle average is performed on the \textit{unknown} classes $C_d^*$. As its name suggests, BAUS computes the balanced accuracy while considering only truly unknown individuals.\\

\noindent\textbf{Combined metrics:}
To evaluate overall performance with a single score, we combine the closed-set (BAKS) and open-set (BAUS) metrics using their geometric mean: 
\begin{equation}
\label{eq:normalized} 
    \text{normalized accuracy} = \sqrt{\operatorname{BAKS}\cdot\operatorname{BAUS}}.     
\end{equation} 
The geometric mean is preferred over the arithmetic mean, as it penalizes extreme imbalances, e.g., a trivial classifier predicting only new individuals yields BAKS = 0\% and BAUS = 100\%, resulting in a misleading 50\% arithmetic mean but a more appropriate geometric mean of 0\%.

\begin{table}[t]
\centering
\caption{\textbf{Closed-set performance on WildlifeReID-10k.} Top-1 and Top-5 accuracy and corresponding BAKS for different architectures and training methods. All metrics reported in \%.}
\begin{tabular}{clcccc@{}}
\toprule
 & \textbf{Architecture}  & \textit{Input}   & \textbf{Top1} & \textbf{Top5} & \textbf{BAKS} \\
\midrule
\multirow{4}{*}{\rotatebox{90}{ArcFace}} 
& ConvNeXt-Base    & 224$^2$ & 66.4 & 77.3 & 64.0 \\
& EfficientNet-B3  & 300$^2$ & 70.5 & 80.6 & 67.3 \\
& ViT-Base/p16     & 224$^2$ & \textbf{79.2} & \textbf{88.9} & \textbf{76.9} \\
& Swin-Base/p4w7   & 224$^2$ & 73.2 & 84.9 & 70.5 \\
\midrule
\multirow{4}{*}{\rotatebox{90}{CCE}} 
& ConvNeXt-Base    & 224$^2$ & 81.1 & 90.3 & 78.5 \\
& EfficientNet-B3  & 300$^2$ & 77.8 & 88.5 & 75.1 \\
& ViT-Base/p16     & 224$^2$ & 78.3 & 88.7 & 75.8 \\
& Swin-Base/p4w7   & 224$^2$ & \textbf{81.5} & \textbf{90.4} & \textbf{79.1} \\
\bottomrule
\end{tabular}
\label{tab:closedset}
\vspace{-2mm}
\end{table}

\section{Results}
This section presents an analysis of the performance of our closed-set and open-set baselines, followed by an evaluation of the MiewID foundational model. \\

\noindent{\textbf{Closed-set baselines:}}
Table~\ref{tab:closedset} compares the closed-set identification performance of various architectures trained with ArcFace and cross-entropy losses. Across all metrics Top-1, Top-5, and BAKS—models trained with cross-entropy consistently outperform their ArcFace counterparts, except ViT, which also achieves the highest performance among ArcFace models with a Top-1 accuracy of 79.2\% and a BAKS of 76.9\%. However, the best overall results are obtained with cross-entropy trained Swin-Base, reaching 81.5\% Top-1 accuracy, 90.4\% Top-5, and a BAKS of 79.1\%. \\

\begin{table}[t]
\centering
\caption{\textbf{Open-set performance on WildlifeReID-10k.} Various baseline architectures and scoring methods. AUROC and TNR@95TPR are reported in \%.}
\begin{tabular}{@{}lcccc@{}}
\toprule
\textbf{Architecture} & \textbf{Method} & \textit{Input} & \textbf{AUC} & \textbf{TNR@95} \\
\midrule
 ConvNeXt-Base      & \multirow{4}{*}{MSP} &224$^2$ & 82.0 & 41.2 \\
 EfficientNet-B3    &  & 300$^2$ & 81.2 & 36.3 \\
 ViT-Base/p16       &  & 224$^2$ & 81.0 & 39.2 \\
 Swin-Base/p4w7     &  & 224$^2$ & \textbf{82.9} & \textbf{44.4} \\
\midrule
 ConvNeXt-Base      & \multirow{4}{*}{MLS} & 224$^2$ & 82.2 & 41.9 \\
 EfficientNet-B3    &  & 300$^2$ & 71.9 & 23.2 \\
 ViT-Base/p16       &  & 224$^2$ & 81.5 & 39.7 \\
 Swin-Base/p4w7     &  & 224$^2$ & \textbf{83.2} & \textbf{45.0}  \\
\midrule
 ConvNeXt-Base      & \multirow{4}{*}{NMS} & 224$^2$ & 75.1 & 23.7 \\
 EfficientNet-B3    &  & 300$^2$ & 76.7 & 32.6 \\
 ViT-Base/p16       &  & 224$^2$ & \textbf{81.8} & \textbf{38.4} \\
 Swin-Base/p4w7     &  & 224$^2$ & 76.7 & 33.0 \\
\bottomrule
\end{tabular}
\label{tab:openset}
\vspace{-2mm}
\end{table}

\noindent{\textbf{Open-set baselines:}}
Consistent with the closed-set results, methods based on cross-entropy (MSP and MLS) generally outperform MSP, which is based on ArcFace. Among the cross-entropy approaches, MLS achieves the best results, with Swin reaching the highest AUROC (83.2\%) and TNR@95 (45.0\%). MSP also performs well with Swin (82.9\% AUROC, 44.4\% TNR). NMS lags behind across most architectures, except ViT. More detailed description of open-set results is in Table~\ref{tab:openset}. \\

\noindent{\textbf{MiewUI performance:}}
MiewID demonstrates strong performance across diverse datasets, but the evaluation split significantly impacts the results. As shown in \cref{fig:miewid} and detailed in \cref{table:miewid}, random splits can lead to overly optimistic outcomes, overestimating open-set performance by up to 12.3\% on datasets with timestamps, due to data leakage. 

In contrast, time- and similarity-aware splits introduce a higher difficulty, resulting in lower but more trustworthy performance metrics. These findings reinforce the importance of a split strategy for fair and meaningful evaluation.

We further analyze model behavior under varying thresholds $t_{\rm new}$ for identifying new individuals (see \Cref{fig:normalized}). As expected, a high threshold favors BAUS while suppressing BAKS, and the reverse holds for low thresholds. The normalized accuracy, which balances the two, peaks at 65.1\% and remains stable between 0.6 and 0.7; highlighting this range as a robust operating point for open-set identification.

\cref{fig:miewid} shows the performance of MiewID-msv3: on 9 datasets with timestamps, solid lines compare the time-aware split (orange) with the random split (blue); on 28 datasets without timestamps, dotted lines compare the similarity-aware split (orange) with the random split (blue).

\begin{table}[!ht]
\caption{Performance on MiewID for the closed-set (BAKS) and open-set (area under BAKS-BAUS curve) settings.}
\label{table:miewid}
\centering
\begin{tabular}{@{}lcccc@{}}\toprule
& \multicolumn{2}{c}{\textbf{timestamps}} & \multicolumn{2}{c@{}}{\textbf{no timestamps}} \\ \cmidrule(lr){2-3} \cmidrule(lr){4-5}
\textbf{Setting} & random & time-aware & random & sim-aware \\\midrule
\textit{Closed-set} & 81.5\% & 69.2\% & 80.8\% & 79.3\% \\
\textit{Open-set} & 72.6\% & 55.9\% & 66.1\% & 63.6\% \\
\bottomrule
\end{tabular}
\end{table}

\begin{figure}[!t]
\centering
\begin{tikzpicture}
\begin{axis}[
    height=4.0cm,
    xmin=0, xmax=0.84, ymin=0, ymax=1,
    scale only axis,
    xticklabel style={
        /pgf/number format/fixed,
        /pgf/number format/precision=3,
    },
    scaled ticks=false,
    legend cell align={left},
    width=0.4\textwidth,
    xlabel={BAKS},
    ylabel style={yshift=-3mm},
    ylabel={BAUS},
    legend pos={south west},
    grid={major},
    ]
\addplot [lineB] table[x index=0, y index=1] {\tabTime}; \addlegendentry{timestamps, time-aware}
\addplot [lineA] table[x index=4, y index=5] {\tabTime}; \addlegendentry{timestamps, random}
\addplot [lineD] table[x index=0, y index=1] {\tabSimilarity}; \addlegendentry{no timestamps, sim-aware}
\addplot [lineC] table[x index=4, y index=5] {\tabSimilarity}; \addlegendentry{no timestamps, random}
\end{axis}
\end{tikzpicture}
\caption{Performance difference between the proposed time-aware or similarity-aware split (orange) and the random split (blue) for datasets with timestamps (full lines) and without timestamps (dotted lines). Each point at a curve corresponds to a different threshold $t_{\rm new}$. This figure suggests that the random split artificially inflates the performance of methods.}
\label{fig:miewid}
\end{figure}

\begin{figure}[!t]
\centering
\begin{tikzpicture}
\begin{axis}[
    height=4cm,
    width=0.4\textwidth,
    xmin=0,
    xmax=1,
    ymin=0,
    ymax=1,
    scale only axis,
    ylabel style={
        yshift=-1mm
    },
    xlabel={Threshold $t_{\rm new}$},
    ylabel={Accuracy},
    ylabel style={yshift=-3mm},
    legend style={at={(axis cs:0.02,0.25)},anchor=south west},
    legend cell align={left},
    grid={major},
]
\addplot [smooth, very thick, blue] table[x index=0, y index=1] {\tabAccuracy}; \addlegendentry{BAKS}
\addplot [smooth, very thick, red] table[x index=0, y index=2] {\tabAccuracy}; \addlegendentry{BAUS}
\addplot [smooth, very thick, orange] table[x index=0, y index=3] {\tabAccuracy}; \addlegendentry{normalized}
\end{axis}
\end{tikzpicture}
\caption{Performance of MiewID on the whole dataset for the open-set setting. The normalized accuracy is based on the threshold $t_{\rm new}$ for predicting new individual. The geometric mean in \eqref{eq:normalized} forces the normalized accuracy to be zero whenever BAKS or BAUS are zero.}
\label{fig:normalized}
\end{figure}

%% file: sections/conclusion.tex
\section{Conclusions}

We introduced WildlifeReID-10k, the largest public dataset for animal re-identification, containing over 140,000 images of 10,772 individuals across 33 species. Unlike previous collections, it brings together a wide range of curated and standardized data in a single benchmark, with open-source tools for dataset creation, analysis, and evaluation.

A key contribution of this work is the introduction of realistic evaluation splits. We showed that commonly used random splits cause significant train-to-test leakage, leading to inflated performance. To address this, we proposed time-aware and similarity-aware splitting that better reflects the challenges of real-world wildlife monitoring. These are integrated into the benchmark by default, enabling fair and reproducible evaluation.

Building on the \href{https://github.com/WildlifeDatasets/wildlife-datasets}{WildlifeDatasets} framework, we made several improvements:
(i) 17 datasets were newly added or updated,
(ii) low-quality datasets were removed and others significantly reduced in size,
(iii) the total dataset size was reduced from 178GB to 26GB by removing redundant files and applying bounding boxes,
(iv) multiple labeling errors were corrected, and
(v) hosting the dataset on Kaggle removed the need for manual handling of individual datasets.

We believe WildlifeReID-10k will set a new standard in the field, offering a reliable foundation for training, testing, and comparing animal re-ID models across species. We hope it accelerates progress in both computer vision research and ecological applications, and invite the community to use, contribute to, and build upon this resource. \\

\noindent\textbf{Limitations}:
While WildlifeReID-10k offers the most comprehensive and curated benchmark to date, there are still limitations. First, despite covering 33 species, many taxa remain underrepresented, especially amphibians and reptiles. Second, some datasets include only a few images per individual, which can make learning identity-specific features more difficult. Third, the similarity-aware split relies on visual clustering using pre-trained models, which may not always reflect true encounter boundaries. Finally, although we applied extensive cleaning, some label noise may persist due to the varied quality of the original datasets.

\section*{Acknowldgement} 
This research has been supported by the Ministry of Education, Youth and Sports of the Czech Republic under project SGS-2024-017.